# NATURAL LANGUAGE GENERATION FOR ELECTRONIC HEALTH RECORDS

**Scott Lee[1]**  yle4@cdc.gov

[1]CENTERS FOR DISEASE CONTROL AND PREVENTION

## Abstract

A variety of methods existing for generating synthetic electronic health records (EHRs), but they are not capable of generating unstructured text, like emergency department (ED) chief complaints, history of present illness or progress notes. Here, we use the encoder-decoder model, a deep learning algorithm that features in many contemporary machine translation systems, to generate synthetic chief complaints from discrete variables in EHRs, like age group, gender, and discharge diagnosis. After being trained end-to-end on authentic records, the model can generate realistic chief complaint text that preserves much of the epidemiological information in the original data. As a side effect of the model's optimization goal, these synthetic chief complaints are also free of relatively uncommon abbreviation and misspellings, and they include none of the personally-identifiable information (PII) that was in the training data, suggesting it may be used to support the de-identification of text in EHRs. When combined with algorithms like generative adversarial networks (GANs), our model could be used to generate fully-synthetic EHRs, facilitating data sharing between healthcare providers and researchers and improving our ability to develop machine learning methods tailored to the information in healthcare data.

## 1 Introduction

The wide adoption of electronic health record (EHR) systems has led to the creation of large amounts of healthcare data. Although these data are primarily used to improve patient outcomes and streamline the delivery of care (healthit.gov), they have a number of important secondary uses, including medical research and public health surveillance. Because they contain personally identifiable patient information, however, much of which is protected under the Health Insurance Portability and Accountability Act (HIPAA), these data are often difficult for providers to share with investigators outside their organizations, limiting their feasibility for use in research. In order to comply with these regulations, statistical techniques are often used to anonymize, or de-identify, the data, e.g. by quantizing continuous variables or by aggregating discrete variables at geographic levels large enough to prevent patient re-identification. While these techniques work, they can be resource-intensive to apply, especially when the data contain free text; which often contains personally-identifiable information (PII) and may need to be reviewed manually before sharing to ensure patient anonymity is preserved.

Recent advances in machine learning have allowed researchers to take a different approach to de-identification by generating synthetic EHR data entirely from scratch. Notably, Choi et al. (2017) are able to synthesize EHRs using a deep learning model called a generative adversarial network (GAN) (Goodfellow et al. 2014). By training one neural network to generate fake records and another to discriminate those fakes from the real records, the model is able to learn the distribution of both count- and binary-valued variables in the EHRs, which can then be used to produce patient-level records that preserve the analytic properties of the data without sacrificing patient privacy. Unfortunately, the model does not generate free-text associated with the records, which limits its application to EHR datasets where such information would be of interest to secondary users. In public health, for instance, the chief complaint and triage note fields of emergency department (ED) visit records are used for syndromic surveillance (Lall et al. 2017; Thomas et al. 2018), and pathology reports are used for cancer surveillance (Ryerson & Massetti 2017). In these cases, automated redaction tools are of little use—since models like Choi et al's GAN can only generate discrete data, there would be nothing for an algorithm to de-identify--and providers interested in sharing them to support research must revert to de-identifying the original datasets, which may be time-consuming and costly.

In this paper, we explore the use of encoder-decoder models, a kind of deep learning algorithm, to generate natural language text for EHRs, filling this gap and increasing the feasibility of using generative models like Choi et al.'s GAN to create high-quality healthcare datasets for secondary uses. Like their name implies, encoder-decoder models comprise 2 (traditionally recurrent) neural networks: one that encodes the input sequence as a single dense vector, and one that decodes this vector into the target sequence. These models have enjoyed great success in machine translation (Bahdanau, Cho, & Bengio 2014; Cho et al. 2014), since they can learn to produce high-quality translations without the need for handcrafted features or extensive rule-based grammar. In some cases, they can even translate between language pairs not seen during training (Johnson et al. 2016), a fairly remarkable result. Because of their flexible design, encoder-decoder models have also been applied to problems in natural language generation. For example, by changing the encoder from a Recurrent Neural Network (RNN_ designed to process text to one designed to process acoustic signals, the model can be trained to perform speech recognition (Chan et al. 2015). Likewise, by changing the encoder to a convolutional neural network (CNN) designed to process images, the model can be trained to caption images (Xu et al. 2015; Vinyals et al. 2015). Because of this success in modeling the relationships between diverse kinds of non-text data and natural language, we adopt this framework to generate free-text data in EHRs.

## 2 Methods
*2.1 Data structure and preprocessing*
Our dataset comprises approximately 5.8 million de-identified emergency department (ED) visit records provided by the New York City Department of Health and Mental Hygiene (NYC DOHMH). The records were collected between January 2016 and August 2017, and they include non-text fields, like age, gender, and discharge diagnosis code, as well as free-text entries for discharge diagnosis and chief complaint. In this case, we choose to focus on the chief complaints, which are a good candidate for testing the encoder-decoder model of natural language generation because, like image captions, they are often short in length and composed of words from a relatively limited vocabulary.

Table 1 shows the variables from which we generate our synthetic chief complaints. We include age group, gender, and discharge diagnosis code (HCUP 2009) for their clinical relevance; and mode of arrival, disposition, hospital code, month, and year for the extra situational information they may provide about the encounters. Following Choi et al. 2017, we recode the variables as integers, as shown in column 3 of Table 1, and we expand them to sparse format so that a single observed value of a particular variable takes the form of a binary vector, where each entry of the vector is 0 except for those corresponding to the indices of recoded values; these are 1. In most cases this vector is one-hot, i.e. it has only a single non-zero entry, but because ED patients are often assigned more than one ICD code during a single visit, discharge diagnosis is often not. Concatenating the vectors for each variable gives us a sparse representation of the entire visit, which we then use as input to our text generation model. Although most of the records were complete, approximately 735,000 (15%) in our final dataset were missing discharge diagnoses, and approximately 435,000 (9%) were missing values for disposition. In these cases, the missing values were coded as all-zero vectors before being concatenated with the vectors for the other variables.

| Variable | Original values | Coded values |
| --- | --- | --- |
| Age group | 0 through 110+ in 5-year (inclusive) increments, e.g. 5-9, 10-14, and 20-24 | [0, 22] |
| Gender | M, F, and 4 other categories including non-binary genders | [0, 5] |
| Mode of arrival | 'ambulance', 'car', 'helicopter', 'missing', 'on foot', 'public transportation', 'unknown', AND 'other | [0, 7] |
| Hospital code | 44 3-digit alphanumeric codes | [0, 43] |
| Disposition (without transfer) | 'outpatient admitted as an inpatient to this hospital', 'routine discharge','discharged to home', 'left against medical advice', 'still patient', 'deceased', 'hospice - medical facility', 'hospice - home', 'deceased in medical facility', 'deceased at home', 'deceased place unknown' AND 'unknown' | [0, 11] |
| Disposition (with transfer) | 'transferred to' + {'critical access hospital', 'intermediate care facility', 'long-term care facility', 'nursing facility', 'psychiatric facility', 'rehabilitation center', 'short-term general hospital', OR 'other facility'} | [0, 7] |
| Month | January through December | [0, 11] |
| Year | 2016 and 2017 | [0, 1] |
| Diagnosis code | ICD-9/10 diagnosis codes converted to HCUP CCS code | [0, 283] |

Table 1. Discrete variables in our dataset. The first column shows the variable names, the second column a description of their unique original values, and the third column a bracketed set indicating the range of those values after being recoded during preprocessing.

To preprocess the chief complaints, we begin by removing records containing words with a frequency of less than 10; this is primarily a way to reduce the size of the word embedding matrix in our Long Short

Term Memory (LSTM) network (Hochreiter & Schmidhuber 1997), reducing the model's computational complexity, but is also serves to remove infrequent abbreviations and misspellings. Similarly, we also remove records with chief complaints longer than 18 words, the 95th percentile for the corpus in terms of length, which prevents the LSTM from having to keep track of especially long-range dependencies in the training sequences. Following these 2 steps, we convert all the text to lowercase, and we append special start-of-sequence and end-of-sequence strings to each chief complaint to facilitate text generation during inference. Finally, we vectorize the text, convert each sentence to a sequence of vocabulary indices, and pad sequences shorter than 18 words with zeros (these are masked by the embedding layer of our LSTM).

After preprocessing, we are left with approximately 4.8 million record-sentence pairs, where each record is a 403-dimensional binary vector $R$, and each chief complaint is a 18-dimensional integer vector $S$. Using a random 75-25 split, we divide these into 3.6 million training pairs and 1.2 million validation pairs. To reduce the time needed to evaluate different sampling procedures during inference, we save only 50,000 of the validation pairs for final testing.

*2.2 Modeling*
2.2.1 Model architecture
Following Vinyals et al. (2015), we formulate our decoder as a single-layer LSTM. To obtain a dense representation of our sparse visit representation $R$, we add a single feedforward layer to the network that compresses the record to the same dimensionality as the LSTM cell; in effect, this layer functions as our encoder. Similarly, we use a word embedding matrix to convert our chief complaints from sequences of integers to sequences of dense vectors, and we use a feedforward layer followed by softmax to convert the output of the LSTM to predicted word probabilities at each time step $t$. For a more detailed description of the model architecture, readers are referred to the supplementary materials.

2.2.2 Training procedures
We pretrain the record embedding layer using an autoencoder, but we do not pretrain the word embeddings. We then train the full model end-to-end in mini-batches of 512 record-sentence pairs, using the Adam algorithm (Kingma & Ba 2014) with a learning rate of 0.001 for optimization. Training is complete once the loss on the validation set fails to decrease for 2 consecutive epochs.

2.2.3 Inference procedures
Encoder-decoder models typically require different procedures for inference (i.e. language generation) than for training. Here, we generate our synthetic chief complaints by following a 4-step process:

1. Feed a record $R$ and the start-of-sentence token to the LSTM as input
2. Generate probabilities for the next word in the sequence
3. Pick the next word according to the rules of a particular sampling scheme
4. Repeat steps 1 and 2 until generating the end-of-sentence token or reaching the maximum allowable sentence length (in this case, 18 tokens)

In Step 3, the sampling scheme provides a heuristic for choosing the next word in the sequence based on its probability given the previous words. Common sampling schemes are choosing the word with the highest probability at each time step (greedy sampling); choosing a word according to its probability, regardless of whether it is the highest (probabilistic sampling); and choosing a word that makes the current sequence of words the most likely overall (beam search decoding). In probabilistic sampling, a temperature value may be used to "melt" the probabilities at each time step, making the LSTM less sure about each of its guesses, and introducing extra variability in the generated sentence. Beam search uses a different tactic and keeps a running list of the $k$ best sentences at each step; as $k$ grows, so does the number of candidate sentences the search algorithm considers, giving it a better chance of finding one with a high probability given the model's parameters. In image captioning, beam search tends to work best (Vinyals et al. 2017), but we test all 3 methods here for completeness.

*2.3 Evaluation*
2.3.1 Translation metrics
Provided that our model can generate chief complaints that are qualitatively acceptable, our first task in evaluating our model is to determine which of the sampling schemes listed above produced the highest-quality text. One way of performing this evaluation is by using metrics designed to evaluate the quality of computer-generated translations. These typically focus on comparing the $n$-grams in the synthetic text to the $n$-grams in the authentic text and calculating measures of diagnostic accuracy, like sensitivity and positive predictive value (PPV), based on the amount of overlap between the two sets of counts. Two of the most common methods for doing this are the Bilingual Evaluation Understudy (BLEU) (Papineni et al. 2002), a PPV-based measure, and the Recall-Oriented Understudy for Gisting Evaluation (ROUGE) (Lin 2004), a sensitivity-based measure. A detailed discussion of how these and related $n$-gram-based metrics work is beyond the scope of this paper (readers are referred instead to Vedantam, Zitnick, & Parikh 2015, which provides a nice overview), but we note that they are not especially well-suited for evaluating short snippets of text, and so we use an alternative method based on variable-length $n$-grams for this portion of our evaluation.

In addition to our simplified measures of $n$-gram sensitivity and PPV, we use several vector-space methods to measure the quality of our generated text. The first is the Consensus-Based Image Description Evaluation (CIDEr; Vadantam, Zitnick, & Parikh 2015), which measures quality as the average cosine similarity between the term-frequency inverse-document-frequency (TF-IDF) vectors of authentic-synthetic pairs across the corpus. Like BLEU and ROUGE, CIDEr is based on an $n$-gram language model, and so we modify it to allow for variable-length $n$-grams to avoid producing harsh ratings for synthetic chief complaints simply because they or their corresponding authentic complaints are short. We do note, however, that in the case of zero $n$-gram overlap, any $n$-gram-based metric will be 0, which seems undesirable, e.g. when evaluating pairs like 'od'/'overdose' and 'hbp'/'high blood pressure', which are semantically nearly identical. Therefore, as our final measure, we take the cosine similarity of the chief complaints in an embedding space--in this case, the average of their word embeddings--which, at least empirically, is always non-zero and never undefined. We discuss this measure, as well as our modifications to BLEU, ROUGE, and CIDEr, in the supplementary materials.

2.3.2 Epidemiological validity
Because we would like synthetic chief complaints to support various secondary uses, including public health surveillance and research, we evaluate their epidemiological validity using several measures. In the most basic sense, we would like to ensure that there is no obvious discordance between information in the discrete variables in a particular record and the synthetic text our models generates from them. To evaluate this measure, we pick a handful of words corresponding to common conditions for which patients may seek care in an ED, like 'preg' for pregnancy and 'od' for overdose, and compare their distributions in the real and synthetic text fields relative to key demographic variables, like gender and age group. Extending this idea further, we also compare odds ratios for certain words appearing in a chief complaint given these same kinds of characteristics, for example the odds of having a complaint containing the word 'fall' given age for patients over 80 as compared those for patients in their 20s.

Our other main measure of epidemiological validity is to see whether we can predict diagnoses from the synthetic chief complaints as well as we can from the real ones. Although there are many methods for performing this task (see Conway, Dowling, and Chapman 2013 for an overview), we use a kind of RNN called a gated recurrent unit (GRU), which Lee et al. (2018) show to have superior performance to the multinomial naive Bayes classifiers employed by several chief complaint classifiers currently in use (details on model architecture and training procedures are provided in the supplementary materials). After training the model on the record-sentence pairs in our training set, we then generate predicted Clinical Classification Software (CCS) (HCUP 2009) codes for the authentic chief complaints in the test set and evaluate diagnostic accuracy using weighted macro sensitivity, positive predictive value, and F1 score. Finally, we compare these scores to the model's same scores on the synthetic chief complaints generated from the same records to gauge how well they capture the relationship between the text and the diagnosis. Although our primary goal in performing this evaluation is to check the epidemiological validity of the synthetic chief complaints, we also use it as an alternative way to measure the quality of text generated by the different sampling schemes during inference.

2.3.3 PII removal

To protect patient confidentiality, patient names are replaced with a secure hash value before being sent to the NYC DOHMH, and other potentially-sensitive fields are anonymized, e.g. by binning ages into 5-year age groups. Although it is a free-text field, chief complaint is largely free of personal identifiers, and so our data are not optimal for exploring our model's potential to de-identify text fields with more sensitive information, like triage notes. Still, the chief complaints do contain the names of a small number of people peripherally involved in ED admissions process, like police officers and referring physicians, and so we conduct a simple evaluation of this potential by seeing if any of these names appear in the model-generated text.

In order to determine whether the generated text contains any names, we start by building a list of names that appear in the original chief complaints. Rather than doing this manually, we train word vectors on our corpus and then search the embedding space for nearest neighbors to known names. Our process follows these steps:

1. Train word embeddings on the entire corpus of chief complaints (here we use the skipgram implementation of word2vec (Mikolov et al. 2013)
2. Randomly pick a name in the data and retrieve its 100 nearest neighbors in the embedding space
3. Manually check the neighbors for other names and add new ones to the list
4. Repeat Steps 2 and 3 until no new names appear in the 100 nearest neighbors

This iterative process yields a list of 84 unique names, which we take to be relatively complete, though by no means exhaustive. Using greedy sampling, we then generate synthetic chief complaints for the 1.6 million records in our combined validation and test sets, and we compare the number of times any of the 84 names appears with their counts in the authentic chief complaints from the same records. Because greedy sampling chooses the most probable word at each time step during inference, our hypothesis is that none of the names, which are low-frequency in the original corpus and thus relatively improbable, will appear in the synthetic text.

*2.4 Technical Notes*
All neural networks were coded in Keras with the TensorFlow (Abadi et al. 2016) backend and trained on a scientific workstation with a single NVIDIA Titan X GPU. Data management was done in Python using the pandas, NumPy (Walt, Colbert, & Varoquaux 2011), and h5py packages, and diagnostic statistics were produced using scikit-learn (Perdregosa et al. 2011).The code for the main model is available on GitHub.

This analysis was submitted for Human Subjects Review and deemed to be non-research (public health surveillance).

**3 Results**
The encoder-decoder model was able to produce rich, variable, and chief complaints. Table 2 shows synthetic complaints produced using 2 different sampling schemes, along with a selection of the discrete variables from their corresponding records. In all cases, the synthetic text aligns well with the diagnosis. The model was also able to generate novel chief complaints, i.e. those not present in the training data. For example, greedy sampling in the test set yielded 3,597 unique sentences, of which 1,144 were novel.

*3.1 Translation metrics*
Table 3 shows the mean n-gram positive predictive value, sensitivity, F1, CIDEr, and embedding similarity scores for the text generated by each sampling scheme. Like Vinyals et al. (2015; 2017), we find that using larger values of $k$ for the beam search decoder did not improve the quality of the synthetic text. Greedy sampling, which is equivalent to beam search with a $k$ of 1, achieved the highest scores, while probabilistic sampling (t=1.0) performed the worst. Although the scores are not directly comparable, we also note that the range of embedding similarity scores is much narrower than the range of scores for the other measures; we attribute this to the fact that embeddings can reflect semantic similarity between related phrases even when there is no $n$-gram overlap.

| Age | Gender | CCS code | CCS description | Chief complaint | Greedy sample | Beam samples (k=5) |
|---|---|---|---|---|---|---|
| 70-74 | M | 660 | Alcohol-related disorders | 'alcoholic beverage consumption today' | 'pt admits to drinking alcohol' | 'etoh', 'etoh consumption', 'found on street', 'found on the street', 'fond on street admits to drinking' |
| 50-54 | F | 2607 | E Codes: Motor vehicle traffic | 'mv c pt denies complaints' | 'pt was rear ended' | 'mvc', 'mva', 'pt was rear ended', 'pt was in a car', 'pt was rear ended in mvc' |
| 20-24 | F | 128 | Asthma | 'shortness of breath on and off x three week' | 'shortness of breath' | 'wheezing', 'cough', 'chest pain', 'chest tightness', 'shortness of breath' |
| 25-29 | F | 661 | Substance-related disorders | 'od narcan given by ems' | 'found on street' | 'od', 'as per ems', 'as per ems patient was found unresponsive', 'as per ems patient was found on the street', 'as per ems patient was found in the street' |
| 85-89 | F | 205 | Spondylosis; intervertebral disc disorders; other back problems | 'm5440 lumbago with sciatica unspecified side i10 needs dialysis' | 'lower back pain m549' | 'm542 cervicalgia i10', 'back pain m545 low back', 'back pain m545 low back pain', 'back pain m545 low back pain i10', 'back pain m545 low back injury i10' |
| 50-54 | F | 98 | Essential hypertension | 'generalized chest pain and blurred vision x few days no travel see above complaint' | 'elevated bp' | 'elevated bp', 'elevated blood pressure', 'high blood pressure', 'elevated bp no travel', 'elevated blood pressure no travel' |

Table 2. Samples of synthetic chief complaints using greedy sampling and beam search, along with key key variables from the corresponding records. We include the authentic chief complaint for comparison, but we omit from the table hospital code, mode of arrival, disposition, month, and year to increase patient anonymity.

*3.2 Epidemiological validity*

In addition to achieving reasonably high scores on the translation metrics above, the synthetic chief complaints also preserve much of the epidemiological information in the original record-sentence pairs. For example, the abbreviation 'preg' appears in the authentic chief complaints for 134 females but for 0 males, and it appears in the synthetic chief complaints for 44 females but also for 0 males. Similarly, 'overdose' has the exact same distribution by 2 age groups in both the authentic and the synthetic chief complaints, appearing 10 times for patients aged 20 to 24 but 0 times for those between 5 and 9. We take these patterns as rough indications that the model can learn to avoid generating highly improbable word-variable pairs.

|  | ppv | sens | f1 | CIDEr | ES |
|---|---|---|---|---|---|
| Beam (k=3) | 0.3323 | 0.1786 | 0.2118 | 0.2013 | 0.6207 |
| Beam (k=5) | 0.3216 | 0.1581 | 0.1922 | 0.1865 | 0.5981 |
| Beam (k=10) | 0.3190 | 0.1410 | 0.1765 | 0.1748 | 0.5733 |
| Prob (t=0.5) | 0.3148 | 0.2208 | 0.2394 | 0.2186 | 0.6541 |
| Prob (t=1.0) | 0.1805 | 0.1520 | 0.1492 | 0.1410 | 0.5973 |
| Greedy | **0.3608** | **0.2418** | **0.2674** | **0.2458** | **0.6688** |

Table 3. Scores for different sampling schemes on our range of text quality metrics. Simplified positive predictive value (PPV), sensitivity (sens), and F1-scores measure *n*-gram overlap between the authentic and synthetic chief complaints; and CIDEr and the embedding similarity (ES) scores measure their similarity in vector space.

Extending this line of analysis to crude odds ratios (ORs) shows similar, though not identical, effects. To illustrate this point, we examine falls, which are a leading cause of injury among older adults (Burns 2018), and which we expect to differ in distribution by age. The authentic data bear this expectation out: in the authentic CCs, the word 'fall' is nearly 8 times more likely to appear for patients who are over 80 years of age (207 of 2,234 patients reporting) than for those between 20 and 24 years of age (48 of 4,009 patients reporting). We see a similar pattern in the synthetic chief complaints, although the association is much stronger, with the older patients being about 15 times more likely to report a fall than the younger patients (229 vs. 27 patients reporting, respectively). We speculate on reasons for this increase in the discussion--it is not unique to falls and appears to be a direct result of the model's particular optimization goal--but we note here that odds of older patients receiving an actual diagnosis of a fall (CCS code 2603) relative to the younger patients falls between these 2 extremes (OR=12.71).

Finally, we also see that the model preserves (and in some cases, amplifies) the relationships between the CCs and the discharge diagnosis codes themselves. Table 3 shows the sensitivity, PPV, and F1 scores for our chief complaint classifier in predicting CCS code from both the authentic chief complaints in the test set ('Original') as well as the synthetic chief complaints generated by the different sampling schemes. The classifier achieved the highest scores on the synthetic chief complaints generated by greedy sampling, but it performed reasonably well on all the samples, differing by no more than 6.5 percentage points on F1 from its performance on the authentic chief complaints.

*3.3 PII removal*
Following the iterative process described above, we discovered 84 unique physician names in the chief complaints. Two-hundred and twenty-four of the 1.6 million authentic chief complaints in the combined validation and test sets contained any of the names; for the corresponding 1.6 million synthetic chief complaints, this number was 0. In this limited sense, the model was successful in removing PII from the free-text data.

|              | sens   | ppv    | F1     |
|-------------:|--------|--------|--------|
| Original     | 0.4487 | 0.4609 | 0.4192 |
| Beam (k=3)   | 0.4892 | 0.5624 | 0.4436 |
| Beam (k=5)   | 0.4687 | 0.5447 | 0.4275 |
| Beam (k=10)  | 0.4354 | 0.5319 | 0.4001 |
| Prob (t=0.5) | 0.4931 | 0.5432 | 0.4481 |
| Prob (t=1.0) | 0.3859 | 0.4053 | 0.3547 |
| Greedy       | **0.5196** | **0.5839** | **0.4713** |

Table 4. Sensitivity (sens), positive predictive value (PPV), and F1 scores for a chief complaint classifier trained on authentic chief complaints and tested on synthetic chief complaints generated with different sampling schemes.

**4 Discussion**

Because our model is trained to maximize the probability of sentences given the information in their corresponding records, the sampling schemes tend to choose high-frequency (and thus high-probability) words when generating new sentences during inference. This strategy has the benefit of achieving strong scores on our various translation metrics, and it also appears to remove PII from free-text, which is encouraging. Additionally, by removing low-frequency terms, the model effectively denoises the original text in the training data, improving our ability to extract important clinical information from the text. As evidence of the latter, we note again that our chief complaint classifier did better at predicting discharge diagnosis from the synthetic chief complaints than from the authentic ones, improving its F1 score by about 5 percentage points.

Still, this optimization strategy has the notable drawback of reducing some of the linguistic variability in the original text that makes it so useful for research and, most especially, for surveillance. For instance, epidemiologists in syndromic surveillance often look for new words in chief complaints to detect outbreaks or build case definitions (Lall et al. 2017), but because these words would be low-frequency, our model would weed them out, limiting the utility of a dataset featuring its synthetic text in place of the authentic text for active surveillance. Similarly, the model tends to choose the most canonical descriptions of the discrete variables in the data, and so, like the crude OR for 'fall' discussed above, it often amplifies associations between specific words and patient characteristics. This distortion could be problematic under a data-sharing model where EHR providers supplied external collaborators with synthetic data for hypothesis generation or exploratory data analysis before granting them access to the original data, although it may have less of an impact on tertiary research activities, like software development in computational health or machine-learning-assisted disease surveillance.

Clearly, though, our model's design has benefits--namely, it enables the LSTM to generate realistic, albeit somewhat homogenized, descriptions of medical data in EHRs--and it opens up several promising lines of research. One avenue is to apply our text generation model to a set of synthetic records, for example those generated by a GAN (Choi et al. 2017), to see how well the resulting text preserves information from the original authentic records. The primary use for a dataset like this would likely be methodological development, and so it is less important that the text retain rare words that would be useful for certain surveillance activities, like outbreak detection, than that it produce similar relationships between those words and discrete variables like diagnosis code. Another possible next step is to adapt the model so that it can generate more complex, perhaps hierarchical, passages of text. Triage notes, for example, are rich sources of information about patient symptoms during ED encounters, but they tend to be longer and more variable than chief complaints, and so they may be difficult to capture with the basic model architecture proposed here. Like other clinical notes, triage notes may also contain unstructured data not quite meeting the definition of natural language, like date-times, phone numbers, zip codes, and medication dosages. The main challenge here, of course, is that the encoder-decoder model can only learn to generate descriptions of information in the input, and so if these other pieces of information are missing from the EHR, it may fail to produce them in the notes, at least in any meaningful way. This is a significant hurdle, but we are optimistic it can be overcome by making adjustments to the model's architecture.

## 5 Conclusion

Our encoder-decoder model was able to generate realistic, clinically-informed natural language descriptions of discrete variables in emergency department EHRs.

**Disclaimer**

**Acknowledgements**

Thanks go to Ramona Lall, Robert Mathes, and the BCD Syndromic Surveillance Unit at the New York City Department of Health and Mental Hygiene for providing the data for the study; and to Chad Heilig at CDC, who provided valuable feedback on the manuscript.

**Supplemental Materials**

*S1. Model architecture*

Like the image captioning model in Vinyals et al. (2015; 2017), our model is designed to directly maximize the following (log) probability, where $R \in \{0,1\}^{403}$ is a sparse representation of the discrete variables in a health record and $S$ is a sequence of words of length $N$ associated with $R$:

$$\log p(S \mid R) = \sum_{t=0}^{N} \log(S_t \mid R, S_0, \ldots, S_{t-1})$$

The encoder for our model is a simple feedforward neural network that converts the sparse record $R$ into a 128-dimensional dense vector:

$$x_{-1} = W_r R$$

Because this vector is fed to the LSTM before the sequence of words, we denote its timestep as *t*=-1. We convert each word $S_t$ in the sentence $S_t$, $t \in \{0 \ldots N-1\}$ to a 128-dimensional dense vector $x_t$ using a word embedding matrix $W_e$:

$$x_t = W_e S_t, \quad t \in \{0 \ldots N-1\}$$

For our language model, we use a single-layer RNN with a LSTM cell, which like the other layers, is 128-dimensional (biases omitted for the sake of simplicity):

$$i_t = \sigma(W_{ix} x_t + W_{im} m_{t-1})$$
$$f_t = \sigma(W_{fx} x_t + W_{fm} m_{t-1})$$
$$o_t = \sigma(W_{ox} x_t + W_{om} m_{t-1})$$
$$c_t = f_t \odot c_{t-1} + i_t \odot h(W_{cx} x_t + W_{cm} m_{t-1})$$
$$m_t = o_t \odot h(c_t)$$
$$p_{t+1} = \text{Softmax}(m_t)$$

The input, output, and forget gates have sigmoid activations $\sigma$, and the cell state has hyperbolic tangent activations *h*. After setting the initial cell state in the LSTM to 0, we feed it the dense record vector, and then we feed it the word embeddings. Unlike Cho et al. (2014), we only show the LSTM the record once, i.e. we do not concatenate it to the word embeddings at each timestep *t*.

After using an autoencoder to pretrain the weights in the encoder (mini-batch size 256; 15 epochs), we train the full model end-to-end, minimizing categorical cross-entropy loss:

$$L(R, S) = -\sum_{t=1}^{N} \log p_t(S_t)$$

*S2. Explanation of modified n-gram overlap statistics*

Both BLEU-N and ROUGE-N calculate scores for different values of *n*, typically from unigrams (*n*=1) up to four-grams (*n*=4); these scores are averaged to obtain the overall score for a particular sentence, and scores for sentences are averaged obtain a score for the entire corpus. As an example, BLEU calculates modified *n*-gram precision for each value of *n* in a sentence, and then takes their geometric mean as the final score, adjusting by an additional brevity penalty to encourage systems to generate longer snippets of text. There are several known issues with the metric--in particular, that it tends not to correlate as well with human judgments of translation quality as other metrics (Vedantam, Zitnick, & Parikh 2015)--but most important for our study is that it was not designed to evaluate short translations. As an example, we can consider the following pair of chief complaints, the first being authentic and the second synthetic:

> Reference: 'cerebral infarction due to unspecified o lusion or stenosis extremity weakness stroke'
> Candidate: 'altered mental status unspecified cerebral infarction unspecified'

The sentences have some *n*-gram overlap at *n*=1 and *n*=2 and describe similar conditions, but there is no overlap at *n*=3 and *n*=4, and so the BLEU-4 score is 0.0. Smoothing functions address this discontinuity (Chen & Cherry 2014), but in ways that seem ad-hoc and that do not naturally transfer to ROUGE. Another issue with both metrics is that that the maximum value of *n* to consider is fixed, which creates undesirable behavior when the length of either the reference sentence or the candidate sentence is less than *n*. Here, we can consider the following pair of chief complaints, where the authentic complaint is much shorter than the synthetic complaint:

> Reference: 'heat stroke hypertension'
> Candidate: 'biba came in hospital for evaluation'

Because the reference sentence only has 3 words, it impossible for it to contain any of the 3 4-grams present in the candidate, and so evaluating BLEU at *n*=4 again leads to a score of 0. In this particular case, the score would be 0 anyway because there is no overlap at the other levels of *n*, but in general we would like to avoid penalizing a synthetic sentence simply because its corresponding authentic sentence is short. More straightforwardly, we would like our metric to make use of higher-order *n*-gram information when it is available, but only when it is available in both sentences.

To address both these issues, we propose a simple measure of *n*-gram overlap that does not require smoothing and is straightforward to calculate. The measure is calculated as follows:

1. Limit *n* to the minimum length of the 2 sentences, if either is less than *n*
2. List the unique 1-through-*n*-grams for each sentence
3. Calculate overlap (either sensitivity or PPV) for each pair of unique sentence *n*-grams
4. Average the sentence scores to obtain a single corpus score

Step 3 is equivalent to calculating the micro-average of the overlap scores for each *n*-gram level, which we find to be better suited to the variable length of the chief complaints than the weighted macro-average used in BLEU-N. Step 2 performs a similar function to the clipping term in BLEU, but because *n*-gram repetition in the chief complaints is often uninformative (consider e.g. the repetition in 'fever unspecified fever unspecified' and 'emesis febrile seizure fever fever and vomiting and became limp'), we opt to ignore it entirely. This step also means that both measures of overlap are calculated from the same pair of *n*-gram sets, which we find to be both intuitive and appealing.

*S3. Explanation of vector-space similarity metrics*
We begin by noting that CIDEr is undefined when either of the sentences is shorter than the maximum value of *n* under consideration--because the magnitude for one of the TF-IDF vectors will be 0, the product of the magnitudes for both the vectors will also be 0, and cosine similarity will be undefined. We therefore modify this metric in the same was as our simplified measures of *n*-gram overlap described above by allowing *n* to vary according to the length of the sentences being compared.

We also note that BLEU, ROUGE, and CIDEr are all 0 when there is no *n*-gram overlap between sentences. Sometimes this behavior is desirable, but because the chief complaints often comprise only 1 or 2 words, it seems especially harsh (consider, e.g. that 'overdose', and 'od' would receive scores of 0 despite their clear semantic similarity). Our solution here is to take the cosine similarity between dense vector representations of the text rather than those based on *n*-grams; as long as the words in the sentences appeared in the training data for the embeddings, this measure is never undefined, and is almost always non-zero. We represent the sentence embedding as follows, where *x* is the one-hot vector for a single word; $W_e$ is the word embedding matrix; *m* is the number of words in the sentence; and *v* is the average of the word embeddings appearing in the sentence:

$$v = \frac{1}{m} \sum_{i=0}^{m} W_e x_{w_i}$$

We then represent the semantic similarity between the reference sentence *r* and the candidate sentence *c* as the cosine similarity of their sentence embeddings $v_r$ and $v_c$:

$$ES(r,c) = \frac{1}{N} \sum_{i=0}^{N} \frac{v_{r_i} \cdot v_{c_i}}{\| v_{r_i} \| \| v_{c_i} \|}$$

Although we did not pursue this adjustment, we note that it would be possible to combine this metric and CIDEr by applying the unigram TF-IDF weights to the word embedding matrix.

*S4. Technical details about the bidirectional GRU*
We implemented our chief complaint classifier as a bidirectional RNN (Graves & Schmidhuber 2005) with a 200-dimensional word embedding layer, a 100-dimensional GRU as the hidden cell, and a 284-dimensional softmax layer to predict the primary CCS code for each record. With a mini-batch size of 128, the model was trained on authentic chief complaints from the training set until its loss on the validation set did not improve for 2 epochs; this occurred after 15 epochs. We again used Adam for optimization, with the learning rate set to 0.001.